\documentclass[12pt, letterpaper]{article} 

\usepackage[T1]{fontenc} 
\usepackage[english]{babel} 
\usepackage{graphicx} 
\usepackage[margin=1in]{geometry} 
\usepackage{amsmath} 
\usepackage{abstract} 
\usepackage{enumitem}
\usepackage{array}
\usepackage{float}
\usepackage{xcolor}
\usepackage[numbers,sort&compress]{natbib}
\usepackage{setspace}
\usepackage[hidelinks]{hyperref} 

\definecolor{PerspectiveBox}{HTML}{EFEEE6}
\newcounter{perspectivebox}

\title{
\textbf{Life Operators: a self-evolving framework for multiscale life modelling}}

\author{
  Shuo Wang\textsuperscript{1} \qquad Yike Guo\textsuperscript{2}\\[0.75em]
  \small \textsuperscript{1}Digital Medical Research Center, Fudan University, Shanghai, China\\
  \small \textsuperscript{2}Hong Kong University of Science and Technology, Hong Kong, China\\[0.5em]
  \small \texttt{shuowang@fudan.edu.cn} \qquad \texttt{yikeguo@ust.hk}
}

\date{}

\begin{document}

\maketitle

\begin{abstract}

Medical AI is moving beyond recognition towards clinical dialogue and longitudinal prediction. Yet a central question remains: how would a patient's state change under intervention?
Statistical models learn future observations, whereas mechanistic models describe selected processes. Neither provides a common framework for representing patient state, coupling scales or revising failed assumptions. We propose \emph{Life Operators}: task-bounded mappings that define three scientific roles. Perception operators infer task-relevant biological states from multimodal observations, Evolution operators propagate these states under natural or intervention-conditioned dynamics, and Generation operators map them to measurable signals. Each role may be realised by equations, statistical models, neural networks or hybrids.
Bridge operators connect components with different variables, scales and time steps. Selected operators and bridges form task-specific Operator Graphs containing the smallest set of states and mechanisms sufficient for a declared claim.
This modular structure also makes scientific revision localisable. An AI co-scientist may propose changes to states, operators, bridges or graph structure, while independent evidence determines which variants are retained, restricted or retired. Over time, validated components could accumulate into broader multiscale models of the human body and provide a computational foundation for medical artificial superintelligence.

\end{abstract}

\section{Introduction}
\begin{figure}[!t]
    \centering
    \includegraphics[width=\textwidth]{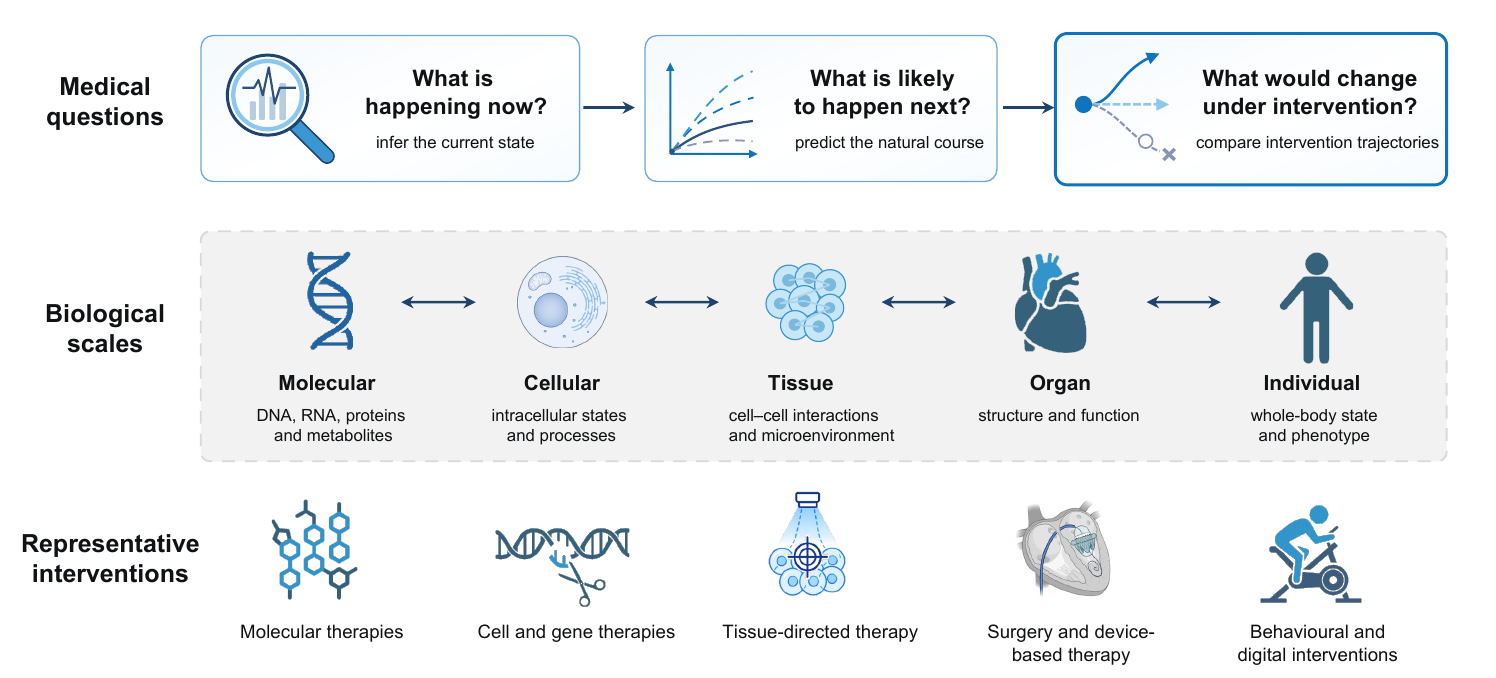}
    \caption{\textbf{Medical questions, biological scales and representative interventions.}
    Three medical questions are aligned with biological organization from molecular to individual scales and representative interventions acting at these levels. Bidirectional arrows indicate interactions across adjacent scales.}
    \label{fig:medical-questions}
\end{figure}

Medicine asks three progressively more demanding questions: What is happening now? What is likely to happen next under prevailing conditions? What would happen under a specified intervention? The first asks which biological states are consistent with the available observations. The second asks how those states are expected to change over time. The third compares action-dependent trajectories. For the same patient, watchful waiting, a drug, ablation or a device may lead to different future states, clinical endpoints and paths between them. A recognition model may address the first question without representing temporal change. A forecasting model may address the second without specifying how an intervention enters the underlying biological process. Addressing the third requires a model in which interventions enter the biological dynamics explicitly. Figure~\ref{fig:medical-questions} relates these questions to biological scales and to representative interventions at different levels of the system.

Data-driven medical AI is beginning to address a broader range of clinical problems. AI-assisted electrocardiography and echocardiography have entered prospective evaluation for detection and measurement within care pathways\cite{Yao2021AI_ECG,He2023EchoAI}. Conversational systems have progressed from diagnostic dialogue to reasoning over multi-visit disease management\cite{Tu2025AMIE,Lievin2026DiseaseManagement}. Longitudinal models learn temporal structure from electronic health records, imaging, age and genetic risk\cite{Shmatko2025Delphi,Urbut2026Aladynoulli,Liu2026Oncoformer}. Together, these developments extend medical AI from present-state recognition towards disease-trajectory modelling.

Predicting a clinical record is not the same as modelling the biological process that generated it. Recorded events reflect both changes in the patient and the process of care. The latter includes testing, treatment assignment, follow-up and coding\cite{Agniel2018EHRBias}. Predicting future records therefore does not, by itself, identify a task-relevant biological state or its transition law. Likewise, conditioning a forecast on an observed treatment label does not establish what would have occurred for the same patient under an alternative treatment. Such models may remain clinically useful without supporting mechanistic interpretation or individual counterfactual claims.

Mechanistic models make a complementary trade-off: they provide explicit variables and structural assumptions, but usually for selected processes and scales.
Cellular electrophysiology, reaction--diffusion models and systems biology have each made selected processes of biological change amenable to quantitative analysis\cite{Hodgkin1952,Turing1952,Kitano2002SystemsBiology}.
Cardiac digital twins further illustrate how anatomical and physiological parameters can be inferred from patient observations and used to parameterise simulations\cite{CorralAcero2020DigitalTwin,Qian2025CardiacTwins}.
Yet fidelity within a local mechanism does not ensure patient specificity or valid composition across scales.
Parameters may be weakly identifiable, and boundary conditions may be incompletely observed. Effective relations may also become unreliable when transferred to another scale or intervention.
The central problem is therefore not whether data-driven or mechanistic models should prevail.
The challenge is instead to make partial models answer one medical question together, even when they use different mathematical forms and sources of evidence. The assumptions of each model must remain explicit.

We propose \emph{Life Operators} as a common language, not another model class. This language specifies what an executable component claims to infer, evolve or generate and states the assumptions and domain of that claim. Perception, Evolution and Generation denote these three scientific roles, while equations, statistical models, neural networks and hybrids are possible implementations. Bridge operators provide explicit mechanisms for composition, and task-specific graphs organise the components required by a medical question. This modular structure provides a basis for evidence-gated revision. Common roles do not imply common algorithms or equal evidential status.

The immediate objective is not a complete digital replica of a person. We instead begin with a declared endpoint, time horizon, set of interventions and available observations. We then construct the smallest state and graph sufficient for that question. Whether independently validated operators and bridges can accumulate into broader models of the human body is an empirical question that must be tested through bounded medical problems. Medical artificial superintelligence is therefore a possible long-term consequence of such accumulation.

\section{Life Operators}

\textbf{Life as a dynamical system.} For a specified medical problem, health and disease may be represented as an open, stochastic, controlled and partially observed dynamical system. A minimal formulation is
\begin{equation}
  s_{t+\Delta t}
  =
  F_{\Delta t}(s_t,u_t,c_t,\xi_t),
  \qquad
  x_t^{(m)}
  \sim
  p\!\left(x_t^{(m)}\mid s_t,a_t^{(m)}\right).
  \label{eq:state}
\end{equation}
Here, $t$ indexes time and $\Delta t$ is the propagation interval. The term $F_{\Delta t}$ is the corresponding state-transition map. The variable $s_t$ denotes the task-relevant biological state, $u_t$ the intervention and $c_t$ the individual and environmental context. The term $\xi_t$ captures unresolved influences and stochasticity. The observation $x_t^{(m)}$, where $m$ indexes modality, may be an image, a physiological signal, a genomic or transcriptomic measurement, single-cell sequencing data, a molecular assay or a clinical record. The acquisition variable $a_t^{(m)}$ describes how the corresponding modality is acquired and recorded. It includes the device, protocol, sampling, detection and reporting processes. When the modality index is omitted below, $x_t$ and $a_t$ denote the corresponding multimodal collections.

The state is not intended to reproduce the complete microphysical configuration of a person or to compress all available data without loss. It is an abstraction chosen for a question. Let $y_{t:t+\tau}$ denote the outcome trajectory of interest over horizon $\tau$. Within a stated domain, an inferred state is sufficient if the observed past adds no material information about the outcome once $s_t$, the allowed inputs and the context are known:
\begin{equation}
  p\!\left(
  y_{t:t+\tau}
  \mid
  s_t,x_{1:t},u_{t:t+\tau},c_{t:t+\tau}
  \right)
  \simeq
  p\!\left(
  y_{t:t+\tau}
  \mid
  s_t,u_{t:t+\tau},c_{t:t+\tau}
\right).
  \label{eq:sufficiency}
\end{equation}

\begin{figure}[!t]
\centering
\includegraphics[width=\textwidth]{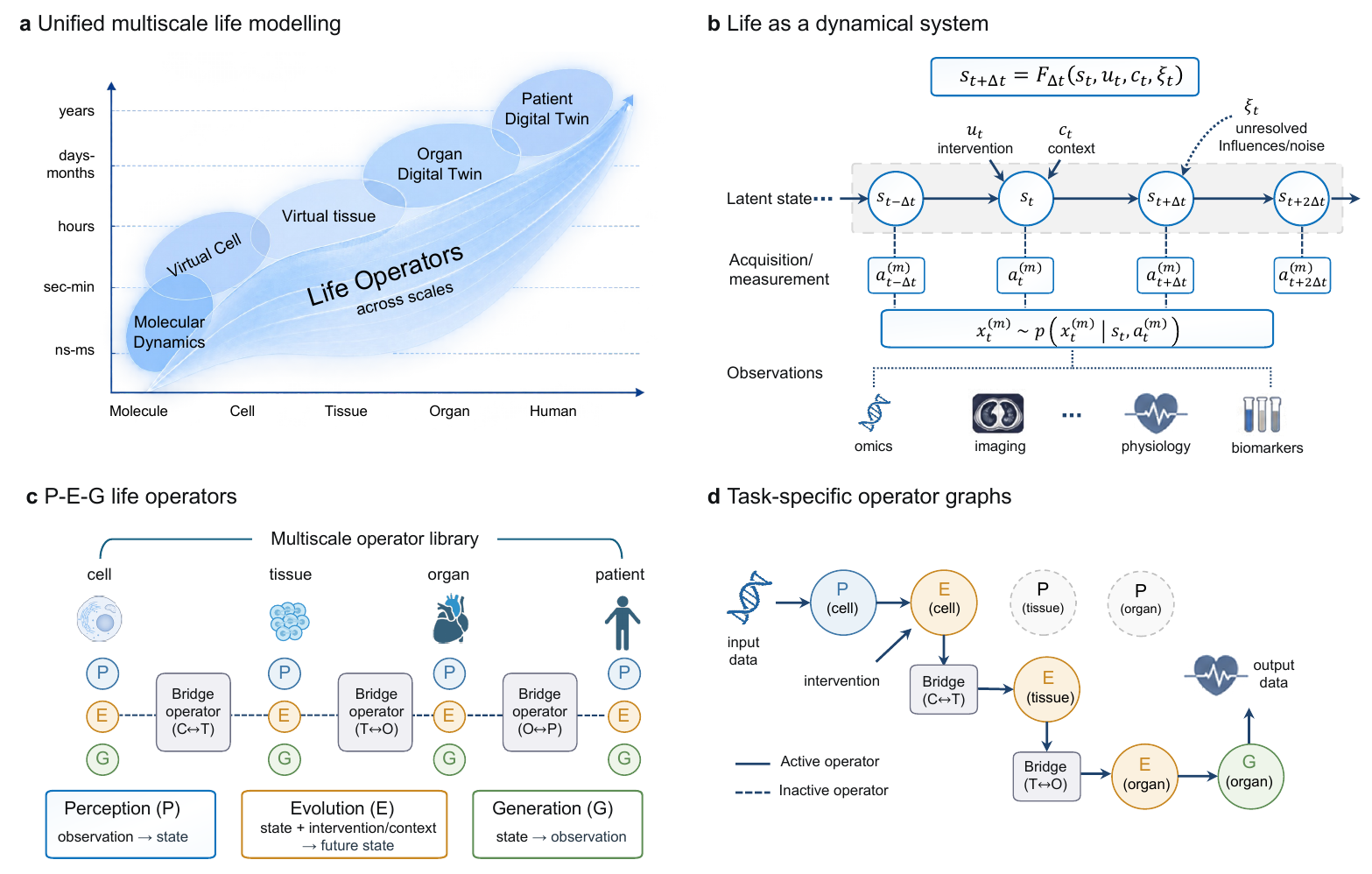}
\caption{\textbf{Life Operators for multiscale modelling.}
\textbf{a}, Models spanning biological scales and characteristic timescales.
\textbf{b}, Life as a partially observed dynamical system shaped by interventions, context and unresolved influences.
\textbf{c}, A multiscale library of Perception, Evolution and Generation operators linked by bridge operators.
\textbf{d}, A task-specific Operator Graph composing selected operators and bridges from input data and intervention to predicted output; dashed nodes are inactive.}
\label{fig:operator-framework}
\end{figure}

Figure~\ref{fig:operator-framework} summarises a progression from multiscale biological models, through dynamical systems and P--E--G operator libraries, to task-specific Operator Graphs.

\textbf{Three operator roles.} The partially observed system exposes three basic computational problems. The latent state must be inferred from observations, propagated under natural or intervention-conditioned dynamics, and related back to measurable signals. We define the corresponding operator classes in parallel:
\begin{align}
\mathcal P &: (x_{1:t},a_{1:t},u_{1:t-1},c_{1:t-1})
\mapsto p(s_t\mid x_{1:t},a_{1:t},u_{1:t-1},c_{1:t-1}), \nonumber\\
\mathcal E_{\Delta t} &: (s_t,u_t,c_t)
\mapsto p(s_{t+\Delta t}\mid s_t,u_t,c_t), \nonumber\\
\mathcal G &: (s_t,a_t)
\mapsto p(x_t\mid s_t,a_t).
\label{eq:operators}
\end{align}
Life Operators are defined by the scientific mappings they claim to perform, not by the algorithms used to implement them. Their implementations may include ordinary or partial differential equations, numerical solvers, statistical transition models and neural networks. Neural operators, learned latent dynamics and hybrid methods may also realise the same role. Mechanistic implementations encode selected structure. Learned components may approximate unresolved relations or accelerate repeated simulation and inversion\cite{Lu2021DeepONet,Kovachki2023NeuralOperator,Karniadakis2021PhysicsInformed}.

The \emph{Perception Operator} maps observations to a posterior distribution over task-relevant biological states. It captures what can be inferred from incomplete observations and the uncertainty of that inference. It therefore avoids treating a single reconstruction as ground truth. This posterior provides the state representation used by the other operators.

The \emph{Evolution Operator} specifies a transition distribution from a current state to a future state over a defined time interval. This distribution is conditional on context and, where relevant, an explicitly defined intervention. It may represent continuous dynamics, stochastic transitions, discrete events or mixtures of these. Within the same dynamical framework, it can describe both natural progression and intervention-conditioned trajectories.

The \emph{Generation Operator} maps a biological state to a distribution of measurable signals. It represents biological signal formation and, where relevant, the processes by which signals are acquired and recorded. In this way, it connects the internal state represented by the model to empirical observations.

\textbf{Inference and uncertainty.} Bayesian inference connects the three roles without forming a fourth operator. For clarity, we use a simplified discrete-time filtering recursion that omits parameters and acquisition variables. In this recursion, Evolution supplies the transition prior, Generation the observation likelihood and Perception the posterior update:
\begin{equation}
 p(s_t\mid x_{1:t},u_{1:t-1},c_{1:t-1}) \propto p(x_t\mid s_t)
 \int p(s_t\mid s_{t-1},u_{t-1},c_{t-1})
 p(s_{t-1}\mid x_{1:t-1},u_{1:t-2},c_{1:t-2})\,\mathrm d s_{t-1}.
 \label{eq:bayes}
\end{equation}
This recursion separates change in the patient from change in knowledge about the patient. Natural variability, measurement error, parameter uncertainty and uncertainty about model form need not be collapsed into a single confidence score.

\section{Operator Graphs}

\textbf{Graphs defined by the question.} Biological processes may be represented by components that differ in variables, scales, units and time steps. A task-specific \emph{Life Operator Graph} (hereafter, Operator Graph) selects and connects the Perception, Evolution and Generation operators required for a medical question. The endpoint, time horizon, interventions and available observations determine which states, operators, bridges and scales the graph requires. For a declared question $Q$, let $\Gamma_Q=(V_Q,E_Q)$. Its nodes $V_Q$ are selected Perception, Evolution and Generation operators. Its edges $E_Q$ are compatible interfaces or bridge operators. The graph is therefore a testable hypothesis about which components are sufficient for the declared claim.

\textbf{Bridges between operators.} The outputs of selected components are not automatically compatible, so each connection requires an explicit transformation. A \emph{bridge operator} transforms the output of one component into the typed input required by another. For a source component $i$ and a receiving component $j$, the minimal relation is
\begin{equation}
r_{i\rightarrow j,t}=\mathcal B_{i\rightarrow j}\!\left(z_t^{(i)}\right),
\label{eq:bridge}
\end{equation}
where $z_t^{(i)}$ is the relevant source state or output and $r_{i\rightarrow j,t}$ is the typed message passed to the receiver. The message may carry an effective parameter, source term, boundary condition or distribution together with its uncertainty. A \emph{Scale Bridge} connects components at different biological scales. Its upward and downward mappings have different meanings and are not assumed to be inverses.

A bridge may be mechanistic, statistical or learned. Regardless of implementation, it must preserve the meaning, units, timing and uncertainty required by the receiving component. Computational compatibility or speed alone does not establish that the resulting Operator Graph is scientifically valid.

Each operator and bridge therefore requires a concise scientific contract. This contract states the component's role, inputs and outputs, variables, units and assumptions. It also records the valid domain, uncertainty, supporting evidence, version and failure conditions. Such contracts make components traceable and comparable, but interface compatibility remains only a prerequisite for composition.

\textbf{Validation before expansion.} Validation must be specific to the receiving component and the declared task. A bridge should preserve the distinctions, uncertainty and intervention responses needed by the receiver. The resulting Operator Graph should be tested against independent observations. If validation fails, the graph may revise the message, narrow its domain or horizon, or activate a finer model.

The graph should expand only when the evidence requires it. Missing observations call for additional measurement or wider uncertainty. A missing scale may require a new bridge or subgraph. Missing evidence for an intervention requires interventional evidence, not additional model complexity. Components that do not improve the declared prediction or its calibration should be removed.

Box~\ref{box:hcm} shows how a bounded question in hypertrophic cardiomyopathy (HCM) can be compiled into explicit states, operators, bridges, outputs and falsification tests.

\begin{center}
\begingroup
\setlength{\fboxsep}{8pt}
\refstepcounter{perspectivebox}\label{box:hcm}%
\colorbox{PerspectiveBox}{%
\begin{minipage}{\dimexpr\textwidth-2\fboxsep\relax}
\footnotesize
\setlength{\parskip}{0.2em}
\setlength{\abovedisplayskip}{4pt}
\setlength{\belowdisplayskip}{4pt}
\textbf{BOX \theperspectivebox}
\par\smallskip\hrule\smallskip
{\normalsize\textbf{An example of building an Operator Graph for drug response in HCM}\par}
\smallskip\hrule\smallskip

This example compiles one prospective test rather than a complete HCM digital twin. For an adult with symptomatic obstructive HCM, it predicts the joint distribution of changes in Valsalva LVOT gradient (VLVOTg) and LVEF at a prespecified follow-up under a declared mavacamten regimen. The cited studies support cell-to-organ simulation, population exposure--response modelling and trial-average effects separately; they do not validate their integration into an individual-response graph\cite{Margara2022HCM,Merali2024Mavacamten,Olivotto2020ExplorerHCM}.

\textbf{Minimal candidate graph.}
\begin{equation*}
\begin{aligned}
x_0
&\xrightarrow{\mathcal P_{\mathrm{base}}}
p(s_0\mid x_0),\\
(u_{0:T},c_{\mathrm{PK}})
&\xrightarrow{\mathcal E_{\mathrm{PK}}}
p(C_{0:T})
\xrightarrow{\mathcal E_{\mathrm{cell}}}
p(T_a,\tau_r),\\
p(T_a,\tau_r)
&\xrightarrow{\mathcal B_{\mathrm{cell}\rightarrow\mathrm{organ}}}
p(\sigma_a),\\
(s_0,\sigma_a,c_{\mathrm{Val}})
&\xrightarrow{\mathcal E_{\mathrm{organ}}}
p(V_{\mathrm{ED}},V_{\mathrm{ES}},v_{\mathrm{LVOT}})
\xrightarrow{\mathcal G_{\mathrm{echo}}}
p(\Delta\mathrm{VLVOTg},\Delta\mathrm{LVEF}).
\end{aligned}
\end{equation*}

\begin{minipage}[t]{0.485\linewidth}
\vspace{0pt}
\textbf{Step 1: Infer the baseline state.}
Inputs $x_0$ include protocolised echocardiography, ventricular and mitral geometry, blood pressure, heart rate, rhythm and treatment history. Perception returns a posterior over $s_0=(g,\ell_0,\kappa,\alpha_0)$: geometry, loading, passive stiffness and baseline contractility. Dose and adherence define $u_{0:T}$; drug-metabolism factors and interacting treatments define $c_{\mathrm{PK}}$; the Valsalva protocol and its loading context define $c_{\mathrm{Val}}$.

\textbf{Step 2: Propagate the intervention.}
Pharmacokinetic Evolution maps the regimen to concentration $C_{0:T}$; cellular Evolution maps concentration to active tension $T_a$ and relaxation $\tau_r$. The Scale Bridge maps these cellular outputs to fibre-oriented active stress $\sigma_a$. Organ Evolution is the key candidate: it combines ventricular geometry, mitral-valve systolic anterior motion, loading and stress to generate pressure--volume dynamics and LVOT flow. It is proposed here, not treated as validated.
\end{minipage}\hfill
\begin{minipage}[t]{0.485\linewidth}
\vspace{0pt}
\textbf{Step 3: Recover a known response.}
Generation applies the echo acquisition protocol $a_{\mathrm{echo}}$ and converts organ states to observables using $\mathrm{LVEF}=1-V_{\mathrm{ES}}/V_{\mathrm{ED}}$ and the simplified Bernoulli relation $\mathrm{VLVOTg}\approx4(v_{\mathrm{LVOT}}\cos\theta)^2$, where $\theta$ is the beam--flow angle, VLVOTg is in mmHg and $v_{\mathrm{LVOT}}$ is in m\,s$^{-1}$. Before individual prediction, the graph should recover the observed population exposure--response relations for VLVOTg and LVEF. This positive control tests intervention response; baseline agreement does not.

\textbf{Step 4: Test each contract.}
Test pharmacokinetics on held-out concentration profiles; cellular dynamics on concentration-matched human force and relaxation data; bridges and organ dynamics on independent deformation, pressure--volume and LVOT-flow observations under unseen loads; and Generation on blinded, protocol-matched echocardiography.

\textbf{Step 5: Freeze, compare and ablate.}
Freeze the graph before external serial evaluation. Compare calibration, prespecified interval coverage and joint predictive score with a simpler exposure--response model, then ablate the cellular and organ components in turn. Retain a scale only if it improves external prediction, calibration or failure localisation.
\end{minipage}

\end{minipage}}
\endgroup
\end{center}

\section{Evidence-Gated Self-Evolution}

\begin{figure}[!t]
\centering
\includegraphics[width=\textwidth]{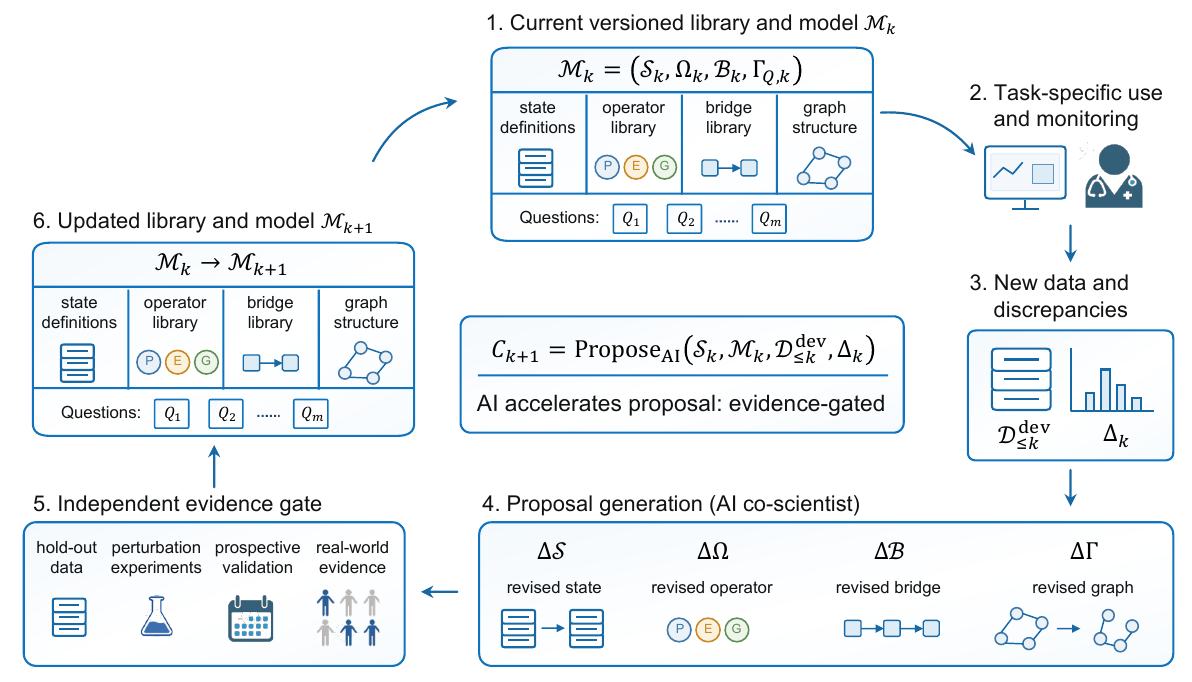}
\caption{\textbf{Evidence-gated self-evolution of a versioned Life Operator model.}
Task-specific use of the current model $\mathcal M_k$ produces new development data and exposes discrepancies. An AI co-scientist proposes revisions to state definitions, operator and bridge libraries, or graph structure. Held-out data, perturbation experiments, prospective validation and real-world evidence gate promotion to the updated model $\mathcal M_{k+1}$. AI accelerates proposal generation; independent evidence determines which changes persist.}
\label{fig:self-evolution}
\end{figure}

\textbf{Modular and versioned revision.} A modular specification represents a bounded biological process and supports systematic revision of the corresponding model. At revision $k$, a versioned operator library is compiled into a task-specific graph. Each component declares its scientific role, interface and valid domain. A failure can therefore be traced to Perception, Evolution, Generation, a bridge or graph composition. The model need not be retrained as a single undifferentiated system. Components can then be added, replaced, restricted or retired while the rest of the graph remains fixed.

Large historical and multimodal datasets can support the learning of reusable operators. New cohorts, measurements, perturbation results and longitudinal records can reveal systematic discrepancies, domain changes and processes missing from the current graph. Explicit operator contracts help localise these gaps to a state, mapping, bridge or connection. Incorporating a new observation into a patient's state posterior remains inference. Self-evolution occurs when accumulated evidence motivates a versioned change in the shared scientific model.

\textbf{Operator discovery through AutoResearch.} Recent AI-enabled research systems show that agents can synthesise literature, generate hypotheses and propose experiments. They can also analyse biological data and refine hypotheses in light of new results\cite{Gottweis2026CoScientist,Ghareeb2026Robin}. In an AutoResearch workflow for Life Operators, agents could inspect residuals and identify missing variables. They could then propose equations or learned mappings, construct bridges and generate code. Such a workflow could move beyond fitting a fixed architecture and propose new Perception, Evolution or Generation operators when existing mappings fail. Each candidate should enter the proposal set with a scientific contract. The contract should specify its inputs, outputs, assumptions, uncertainty, domain and failure conditions. Figure~\ref{fig:self-evolution} summarises this evidence-gated revision cycle. Its proposal and selection steps can be written schematically as
\begin{equation}
\begin{aligned}
\mathcal C_{k+1}
&=
\operatorname{Propose}_{\mathrm{AI}}
\!\left(\mathcal M_k,\mathcal D_{\leq k}^{\mathrm{dev}},\Delta_k\right),\\
\mathcal M_{k+1}
&=
\operatorname{Select}_{\mathcal D_{k+1}^{\mathrm{ind}}}
\!\left(\{\mathcal M_k\}\cup\mathcal C_{k+1}\right),
\qquad
\mathcal M_k=(\mathcal S_k,\Omega_k,\mathcal B_k,\Gamma_{Q,k}).
\end{aligned}
\label{eq:self-evolution}
\end{equation}
Here, $\mathcal S_k$ denotes the state definitions, and $\Omega_k$ denotes the library of Perception, Evolution and Generation operators. The symbols $\mathcal B_k$ and $\Gamma_{Q,k}$ denote the bridge-operator library and graph structure, respectively. Accumulated development data $\mathcal D_{\leq k}^{\mathrm{dev}}$ and discrepancies $\Delta_k$ are used to generate a set of candidate model variants $\mathcal C_{k+1}$. A candidate may change any of these elements rather than merely adjust parameters.

\textbf{Selection by independent evidence.} Independent data or discriminating experiments $\mathcal D_{k+1}^{\mathrm{ind}}$ are used to select among the current model and its proposed revisions. Including $\mathcal M_k$ in the candidate set means that revision is not compulsory. Generating a proposal does not establish its validity. Each candidate should be compared with the current version and simpler alternatives using evidence matched to the intended claim. Held-out observations may support reconstruction or prediction, whereas intervention response may require perturbational, prospective or trial evidence. The same data should not both generate a candidate and provide its final confirmation. When evidence cannot distinguish candidates, the current model should be retained, its domain narrowed or structural uncertainty carried forward.

Accepted operators enter a versioned library and can be recomposed for related medical questions. Repeated cycles of discrepancy detection, candidate generation and evidence-based selection allow the library and its graphs to change as data and knowledge accumulate. This is the intended sense of self-evolution: not unconstrained self-rewriting, but cumulative revision of an explicit, versioned scientific model. 

\section{Biomedical Applications}

With validated Operator Graphs that simulate intervention-conditioned biological trajectories, medicine could move from describing disease to designing ways to change its course. Such a graph would provide an executable environment for designing and comparing candidate interventions across biological scales. It could be revised as new data arrive. The central shift is from asking what is likely to happen to searching how a desired trajectory can be achieved.

\textbf{Personalised therapeutics.} An Operator Graph could be used to design the therapeutic itself for a person, rather than only select from a fixed menu of standard treatments. Molecular profiles, immune state, anatomy, physiology, exposures and prior responses could initialise an individual graph. Candidate molecules, biologics, cell or gene therapies, and combination regimens could then be generated and compared. The graph could also evaluate alternative doses, schedules and delivery strategies in terms of efficacy, toxicity, resistance and long-term outcomes. Personalised mRNA neoantigen cancer vaccines are one example, in which mutation-derived neoantigens and HLA information guide an individual product\cite{Rojas2023RNAVaccine,Sethna2025RNAVaccine}. The graph could then connect antigen presentation and immune dynamics to tumour response and recurrence. The same principle extends to other patient-specific drugs and therapeutic regimens, which could be redesigned as the patient's state changes.

\textbf{Virtual development and trials.} Operator Graphs could connect molecular target engagement, exposure, geometry or energy delivery to cellular response, organ function and clinical outcomes. Large candidate spaces for drugs, devices and procedures could be explored across virtual cohorts representing different biological states. This would support the identification of response and toxicity subgroups, biomarkers, eligibility criteria, endpoints and failure mechanisms. It would also help select the most informative laboratory experiments and clinical comparisons. In silico development would thus become an active design environment. It would focus physical experiments and trials on the candidates and populations most likely to be informative\cite{Viceconti2021InSilico}.

\textbf{Adaptive precision care.} For an individual patient, Perception operators would update the current state from imaging, physiological signals, biomarkers, omics, wearables and clinical records. Alternative drugs, doses, surgery, radiotherapy, ablation, device settings, behavioural interventions or watchful waiting could then be propagated through the same graph. Rather than returning only an endpoint risk, the model could compare intermediate biological changes, time to benefit, adverse effects, recovery paths and future treatment options. Continued observations would reveal deviations from the expected trajectory and identify the next informative measurement. They would also allow prevention, treatment, rehabilitation and follow-up to be adjusted over time. Across patients, the resulting data would expose missing mechanisms and support the self-evolution of the operator library.

These applications would not remove laboratory experiments or clinical trials. They would make them more targeted by identifying which candidates, patient groups and intervention contrasts are most worth testing.

\section{Towards Life Modelling Infrastructure}

Realising these applications requires shared infrastructure and coordinated validation beyond individual operators, task-specific graphs and institutions. It should match claims to evidence, test validity after composition, and support transparent reuse and revision across sites and populations.

\textbf{Evidence matched to claims.} Life Operators should be evaluated against their intended claims. Observation reconstruction and uncertainty calibration require held-out measurements across relevant acquisition conditions. Interpretation of inferred parameters additionally requires identifiability and sensitivity analysis\cite{Gutenkunst2007Sloppy,Villaverde2016Identifiability}. Temporal prediction should use prospectively separated trajectories at declared horizons and across sites and populations. Intervention-response claims require interventional evidence or observational evidence with an explicit causal identification strategy\cite{HernanRobins2020Causal}. Clinical claims require evidence of benefit and safety in the intended workflow. Each operator and graph should declare its intended claim, supporting evidence and unresolved uncertainty.

\textbf{Validity after composition.} A valid operator does not automatically produce a valid Operator Graph. Measurement error, biological variation, parameter uncertainty and bridge design can interact during propagation and feedback. Validation should examine measurable intermediate states, semantic and unit consistency, and uncertainty transport. It should also assess intervention response, long-horizon stability and calibration. Graphs should be compared with simpler predictive, mechanistic or hybrid alternatives at comparable data and computational cost. An additional operator, scale or bridge is justified only when it improves the task, explains a failure or changes an experimental or clinical decision.

\textbf{Infrastructure for reuse and revision.} Reuse requires machine-readable contracts and open, versioned registries for operators and bridges. Each entry should record state semantics, inputs and outputs, units and clocks, assumptions, and valid domains. It should also document provenance, evidence status, uncertainty and known failure conditions. Shared standards, reference implementations and benchmarks would allow the scientific and clinical community to reproduce components, compare alternatives and contribute new operators without rebuilding a graph.

\textbf{International collaboration for validation.} No single institution can assemble dense multiscale and longitudinal measurements throughout the lifespan in diverse populations. Collaboration among population cohorts, hospitals, biobanks and registries can align data elements, intervention definitions and endpoints. When person-level data cannot be pooled, federated analysis can support joint evaluation across sites. Diverse cohorts are needed to test transportability across populations, healthcare systems, devices and clinical practice, and to reveal missing or non-transferable operators. Such collaborations should also support independent testing of AI-proposed candidates and the maintenance of frozen clinical versions, audit trails and rollback procedures.

\textbf{A staged path.} Development can begin with bounded Perception, Evolution and Generation operators whose states can be measured independently. Bridges can then be validated using evidence at both scales. Minimal Operator Graphs can address questions with explicit interventions, measurable intermediate states and prospective endpoints. Shared registries and international cohort networks can provide benchmarks, external validation and evidence-gated reuse. This path begins with tractable questions rather than an attempt to construct a complete digital human. Each task adds tested operators, bridges and graph patterns to a shared system that expands with new data and knowledge.

\bibliographystyle{unsrtnat}
\bibliography{references}

\end{document}